\documentclass{article}

 \usepackage[dblblindworkshop, final]{neurips_2025}
\workshoptitle{Mechanistic Interpretability}

\usepackage[utf8]{inputenc} 
\usepackage[T1]{fontenc}    
\usepackage{hyperref}       
\usepackage{url}            
\usepackage{booktabs}       
\usepackage{amsfonts}       
\usepackage{nicefrac}       
\usepackage{microtype}      
\usepackage{xcolor}         
\usepackage{graphicx}
\usepackage{float}

\title{Vector Arithmetic in Concept and Token Subspaces}

\author{%
Sheridan Feucht \quad Byron Wallace \quad David Bau\thanks{Paper website at \href{https://arithmetic.baulab.info}{\texttt{https://arithmetic.baulab.info}}.} \\
Northeastern University\\
\texttt{\{feucht.s,b.wallace,d.bau\}@northeastern.edu}
}

\begin{document}

\maketitle

\begin{abstract}
  In order to predict the next token, LLMs must represent semantic and surface-level information about the current word. 
  Previous work identified two types of attention heads that disentangle this information: (i) Concept induction heads, which copy word meanings, and (ii) Token induction heads, which copy literal token representations \citep{feucht2025dualroute}. We show that these heads can be used to identify subspaces of model activations that exhibit coherent semantic structure in Llama-2-7b. Specifically, when we transform hidden states using the attention weights of concept heads, we are able to more accurately perform parallelogram arithmetic \citep{mikolov, mikolov2} on the resulting hidden states, e.g., showing that \textit{Athens} -- \textit{Greece} + \textit{China} = \textit{Beijing}. This transformation allows for much higher nearest-neighbor accuracy (80\%) than direct use of raw hidden states (47\%). Analogously, we show that token heads allow for transformations that reveal surface-level word information in hidden states, allowing for operations like \textit{coding} -- \textit{code} + \textit{dance} = \textit{dancing}.
\end{abstract}

\section{Introduction}

Consider how an LLM might model the word \textit{boat}. A boat is a type of vehicle that floats on water, can be powered by sails or engines, and generally carries one or more people. 
But there are many other important facts about this word: It is an English word that is all lowercase; it starts with the letter `b'; it rhymes with (and looks like) \textit{coat}, and it is a singular common noun referring to a broad category. If we wish to analyze the relationship between the word \textit{boat} and the word \textit{water}, we must focus on the semantics of these words, discarding all of the other information that an LLM might encode. On the other hand, if we are trying to find a word that rhymes with \textit{boat}, its meaning may not be particularly helpful to know. 

In the original \texttt{word2vec} paper, Mikolov et al. \cite{mikolov, mikolov2} embed words in a manner that allows for parallelogram-like vector arithmetic: they claim that their embedding space is structured such that \textit{man} is to \textit{woman} as \textit{king} is to \textit{queen}. However, we find that their approach is only somewhat effective for raw Llama-2-7b hidden states \citep{llama2} (Section~\ref{sec:arithmetic}).
We hypothesize that these apparently poor results observed using a naive approach may be attributed to ``interference'' from irrelevant information in model activations.
In other words, we posit that much of the information packed into LLM hidden states has nothing to do with semantics, and that \texttt{word2vec} arithmetic is only effective if performed in a semantic subspace of model activations. 

By using the weights of concept induction heads from \cite{feucht2025dualroute}, we isolate a lower-dimensional space of Llama-2-7b activations for which, e.g., the representation of \texttt{king} - \texttt{man} + \texttt{woman} $\approx$ \texttt{queen}.
We also find that we can use token induction heads to perform parallelogram arithmetic for surface-level tasks, like identifying the first letter in a word, with much higher accuracy than using raw hidden states. This suggests that concept and token induction heads from \cite{feucht2025dualroute} exhibit rich structure in their outputs, operating in subspaces of model activations that represent different facets of words. 

\section{Method: Concept and Token Lens}\label{sec:lenses}

In recent work, Feucht et al. \cite{feucht2025dualroute} identify two types of attention heads responsible for copying text in-context: token induction heads (first described in \cite{elhage2021mathematical}), which copy exact tokens, and concept induction heads, which copy ``fuzzy'' word meanings. 
As these attention heads are responsible for copying previous words seen in-context, their value and output weights can be naturally viewed as transformations that extract semantic and token-level information from any given hidden state. 
Feucht et al. \cite{feucht2025dualroute} use this insight to develop a ``concept lens'' that visualizes semantic information in hidden states. 
We repurpose their approach to derive general concept and token transformations that reveal meaningful structure in hidden states, in the sense that arithmetic in the resultant space accords with intuitive analogies. 

Let $d$ be Llama-2-7b's hidden dimension and $m$ the dimension of a single head. We rely on a key insight from Elhage et al. \cite{elhage2021mathematical}: the value and output projections for a particular head $h$ at layer $l$, $\smash{V_{(l,h)}\in\mathbb{R}^{(m,d)}}$ and $\smash{O_{(l,h)}\in\mathbb{R}^{(d,m)}}$ respectively, are solely responsible for whatever information a head writes into the residual stream. Specifically, they point out that the product of these two matrices $O_{(l,h)}V_{(l,h)}$ is a low-rank $d\times d$ matrix (at most rank $m$) that determines the effect of head $(l,h)$ on the residual stream. In other words, multiplying a hidden state $x_l$ by this matrix extracts whatever information within $x_l$ that this head typically contributes to the residual stream. 

As described in \cite{feucht2025dualroute}, to build a \textit{concept lens} $\smash{L_{C_k} \in \mathbb{R}^{(d,d)}}$ that reads from all of the concept induction head subspaces simultaneously, we combine the weights from the top-$k$ concept induction heads $C_k$. We calculate the sum of the top-$k$ concept OV matrices: 

\begin{equation}\label{eq:concept_lens}
    L_{C_k} = \sum_{(l,h)\in C_k} O_{(l,h)}V_{(l,h)}.
\end{equation}

If all attention heads in $C_k$ are in the same layer, $L_{C_k}x_l$ is mathematically equivalent to taking the sum of the outputs of those attention heads. However, we also allow for summation of heads across layers, which was empirically effective in prior work \citep{todd2024function}, possibly because transformer representations are interchangeable in intermediate layers \citep{lad2025llmsstages}. We can repeat this process using the top-$k$ token induction heads $T_k$ to obtain a \textit{token lens}, which reveals information about the written form of a word. 

\section{Parallelogram Arithmetic}\label{sec:arithmetic}
\subsection{Approach}
We test the assertion made by Mikolov et al. \cite{mikolov} that embeddings should exhibit parallelogram-like structure: in other words, we test whether \texttt{man}~--~\texttt{woman} = \texttt{king} -- \texttt{queen}. Figures \ref{fig:figure1}a and \ref{fig:figure1}b illustrate our approach. We use data from Mikolov et al. \citep{mikolov} and Todd et al. \cite{todd2024function}, which consists of tuples of words in some relation to each other. For every possible pair of tuples, we want to evaluate whether the difference between one tuple is equal to the difference between another; i.e., for (\texttt{Athens}, \texttt{Greece}) and (\texttt{Beijing}, \texttt{China}), we want to evaluate whether \texttt{Athens}~--~\texttt{Greece} = \texttt{Beijing} -- \texttt{China}. In general, we notate this as $a - b = a' - b'$ for a pair of tuples ($a$, $b$) and ($a'$, $b'$). 

To obtain embeddings for each word $w$, we first pass that word through the model in a clean run, obtaining a single vector $w_{\ell}$ by taking its last token representation at a particular layer $\ell$. We then transform this hidden state using some $d\times d$ matrix $L$ to obtain $Lw_\ell$. 
In the \textbf{raw} setting, we do not transform $w_\ell$ at all, so $L=I_d$. In the \textbf{concept} setting, we use $L=L_{C_k}$, in the \textbf{token} setting, we use $L=L_{T_k}$, and as a baseline, we use $L=L_{all}$, which is the sum of \textbf{all} attention head OV matrices. Finally, to see whether $a_\ell-b_\ell = a'_\ell - b'_\ell$ in the subspace mapped to by $L$, we calculate $La_\ell-Lb_\ell+Lb'_\ell$ and evaluate if $La'_\ell$ is its nearest neighbor among all possible words in this task.


Passing a word to a model on its own can be ambiguous, so we choose a prefix for each task that can be prepended to all words in that task (Table~\ref{tab:prefixes}). This sequence is constant across all words for which we perform vector arithmetic. See Appendix~\ref{app:full-nn} for results for all tasks with and without prefixes. 

\vspace{-0.8em}
\subsection{Results}
\begin{figure}
    \centering
    \includegraphics[width=\linewidth]{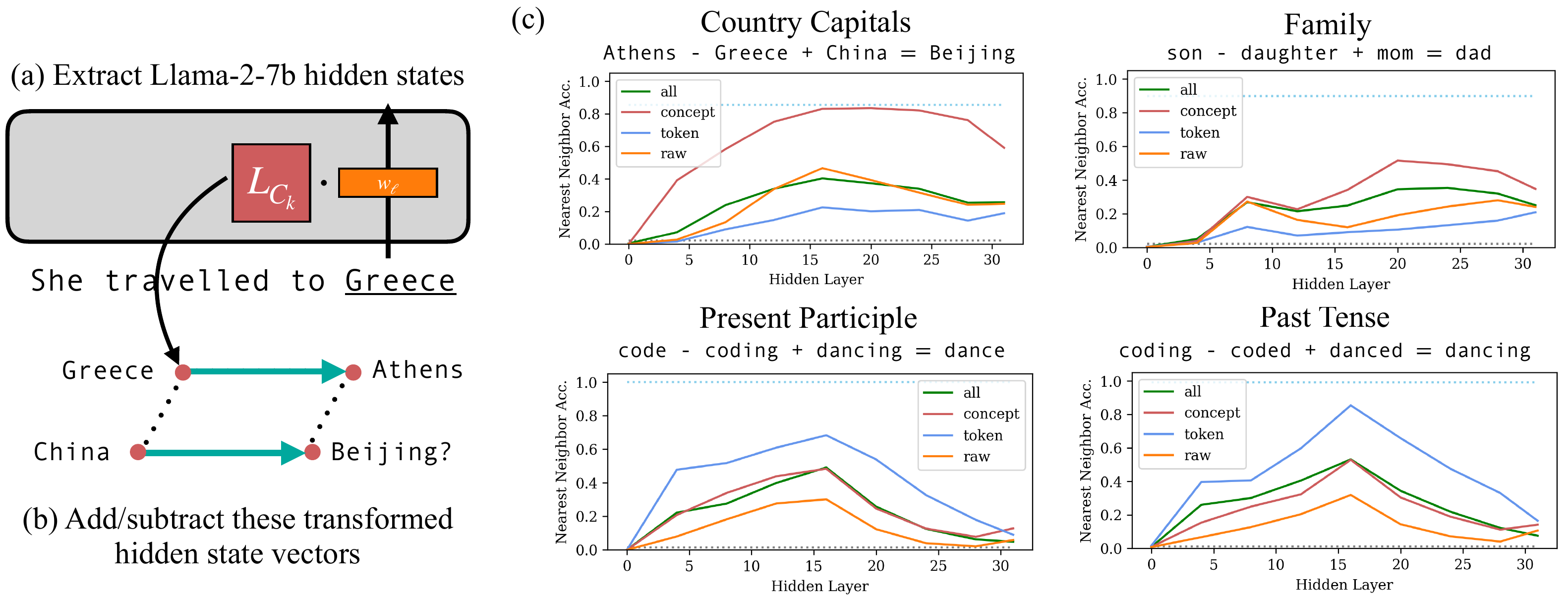}
    \caption{\texttt{word2vec}-style vector arithmetic is more accurate when working in subspaces from \cite{feucht2025dualroute} instead of using raw hidden states. (a) To extract embeddings for a word, we prefix with a constant phrase (e.g. ``She travelled to'') and save the last token representation of the word at a chosen layer $\ell$. To extract conceptual or token information from this vector, we multiply by concept and token lenses $L_{C_k}$ and $L_{T_k}$ respectively (Section~\ref{sec:lenses}). (b) Using a vector from a separate context to represent each word, we measure whether \texttt{Athens} -- \texttt{Greece} + \texttt{China} has \texttt{Beijing} as its top nearest neighbor. (c) For semantic tasks like capital cities and gender-based family words, doing vector arithmetic in the subspace of the top-$k$ concept heads (red) is more effective than using raw hidden states (orange), the top-$k$ token heads (blue), or the sum of all attention head OV matrices (green). On the other hand, the subspace read by the top-$k$ token heads is most effective for grammatical tasks that involve changing the spelling of a word (e.g., \texttt{code} $\rightarrow$ \texttt{coding}). For comparison, dotted gray lines represent random chance, whereas dotted light blue represents Llama-2-7b's 5-shot ICL accuracy for this task. We use $k=80$, as found in \cite{feucht2025dualroute}.}
    \label{fig:figure1}
\end{figure}

Figure~\ref{fig:figure1} shows nearest-neighbor accuracy for selected tasks. While all settings achieve accuracies above random chance (represented by the dotted gray line), concept and token lenses allow for much more accurate vector arithmetic. In the case of capital cities, this arithmetic is just as good as the model's accuracy when asked to complete the task in an ICL setting with 5 shots (light blue dotted line). Oddly, this approach is less effective for tasks that seem simpler, like present participles of verbs. Errors in these cases are difficult to interpret, as the incorrect nearest neighbor is often one of the operands in the original expression. 

\begin{figure}
    \centering
    \includegraphics[width=\linewidth]{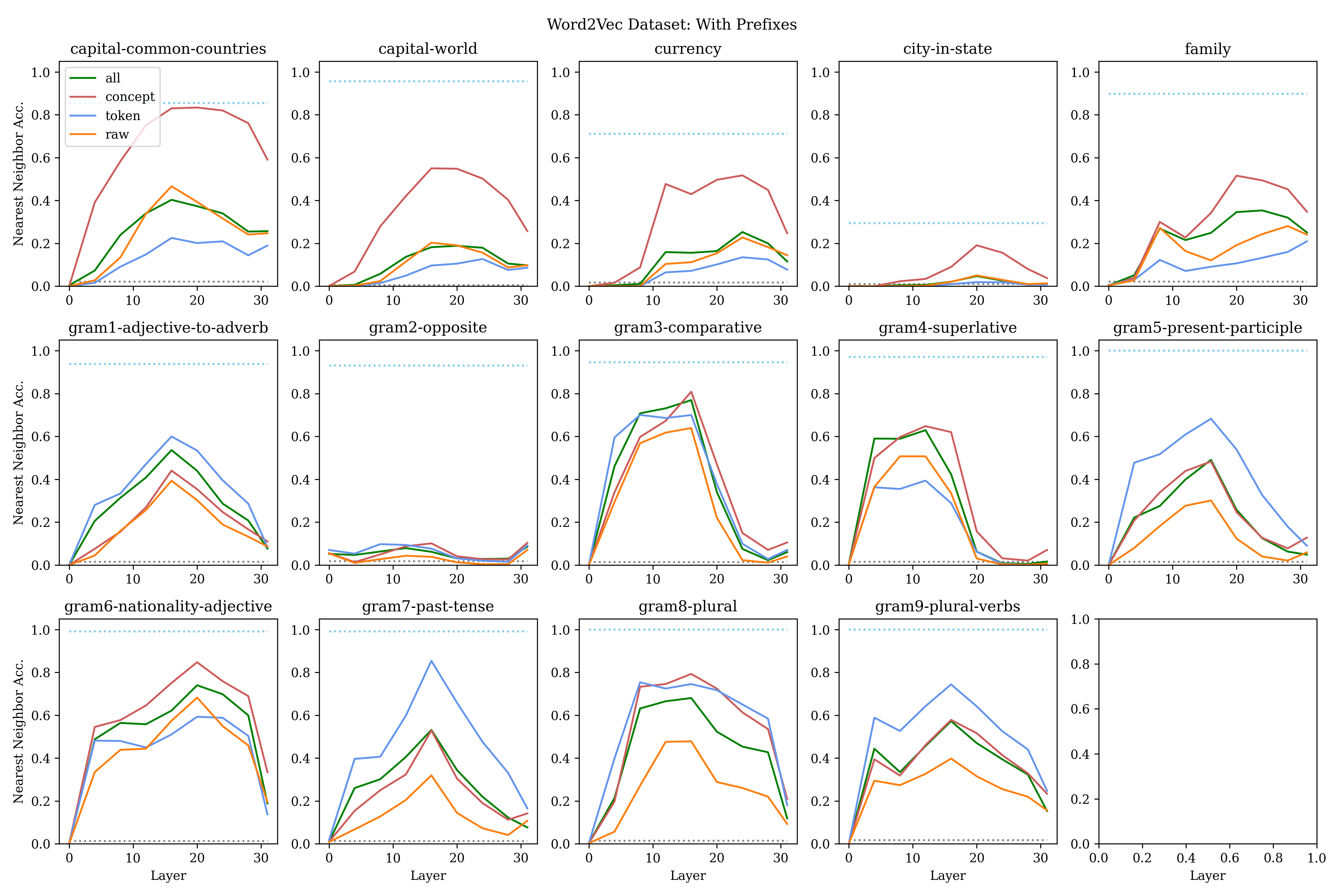}
    \caption{Nearest-neighbor accuracy for all \texttt{word2vec} tasks \citep{mikolov} with prefixes for each task in Table~\ref{tab:prefixes} (Llama-2-7b). Dotted gray lines indicate guessing accuracy (out of all possible neighbors/words in the dataset). Dotted light blue lines indicate 5-shot ICL accuracy for this task, i.e., the best possible performance this model can have for this task. We do not expect high performance for the ``opposite'' task due to its cyclic nature: to represent the concept of ``opposite,'' we need \texttt{possible} -- \texttt{impossible} = \texttt{impossible} -- \texttt{possible}, which is incompatible with parallelogram arithmetic. Targeted subspaces are more effective than using all attention heads for most tasks, except for gram1, gram3, and gram4.}
    \label{fig:word2vec-withprefix}
\end{figure}

Figure~\ref{fig:word2vec-withprefix} shows results for 14 tasks from the original \texttt{word2vec} paper \citep{mikolov}. Concept lens is more effective for semantic tasks, whereas token lens does well for tasks that contain surface-level word variations (e.g., \texttt{quick} $\rightarrow$ \texttt{quickly}). Pluralizing nouns (``gram8-plural'') can be done in both concept space and in token space (by adding `s' to a word), but pluralizing verbs can only be done in token space (``gram9-plural-verbs''), possibly because the latter mostly has to do with verb agreement, not word meaning. See Appendix~\ref{app:full-nn} for more tasks from Todd et al. \cite{todd2024function}.

\vspace{-1em}
\subsection{Effective Rank of Concept and Token Subspaces}

\begin{figure}
    \centering
    \includegraphics[width=\linewidth]{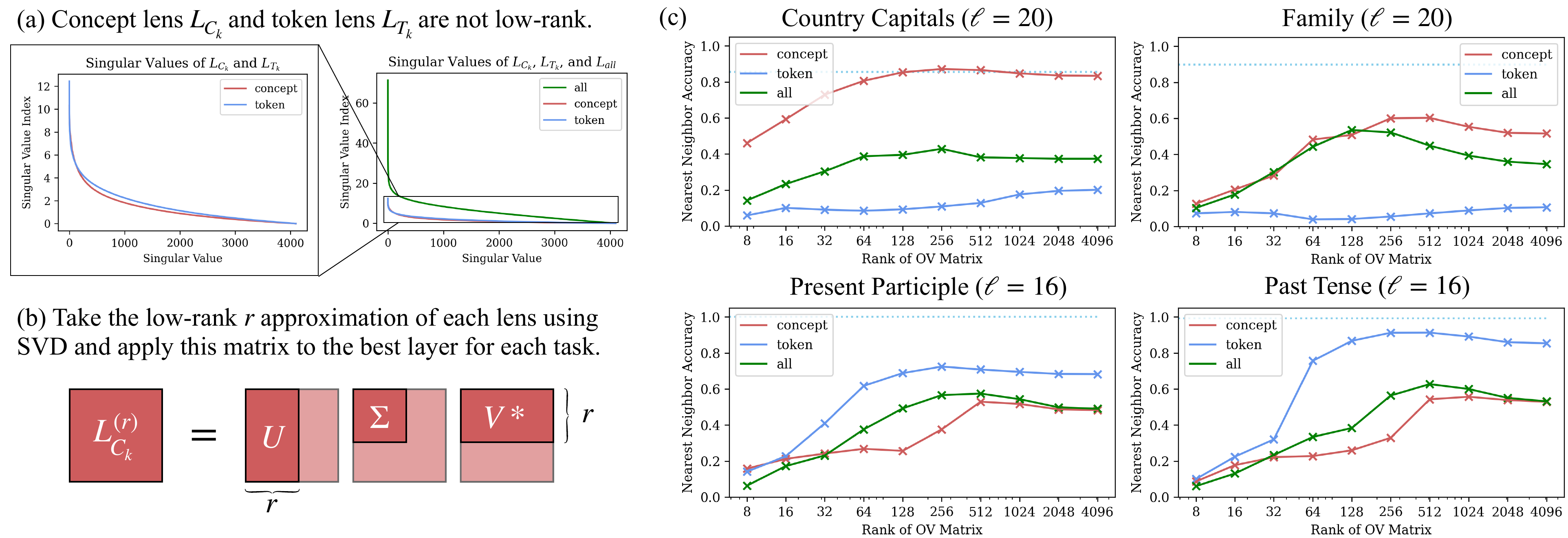}
    \caption{Reducing the rank of $L$ by taking the top-$r$ singular components does not damage nearest-neighbor accuracy. (a) Inspecting the singular values of our concept lens, $L_{C_k}$, and token lens, $L_{T_k}$, these transformations appear to be full-rank. (b) Regardless, we take $r$-rank approximations of these transformations by setting all singular values after the top-$r$ values to zero. (c) We choose the best layer for each task from Figure~\ref{fig:figure1} and reduce the rank of every $L$ in this way. Performance is maintained for ranks as low as $r=256$. Note that values for $r=4096$ are the same as results from Figure~\ref{fig:figure1}.}
    \label{fig:lowrank}
\end{figure}

Although the OV matrix for a single attention head is at most rank $m$ with $m<d$, our transformations $L_{C_k}$ and $L_{T_k}$ are full-rank when $k=80$, as shown empirically in Figure~\ref{fig:lowrank}a.
This means that our transformations for Figure~\ref{fig:figure1} do not actually project activations onto a strict concept or token subspace. However, we hypothesize that we do not need to use all $d$ dimensions to perform vector arithmetic for these tasks. To test this, we set all singular values below the top-$r$ values to zero for $L_{C_k}$, $L_{T_k}$, and $L_{all}$, sweeping across values of $r$ (Figure~\ref{fig:lowrank}b). We choose the best layer for each task from Figure~\ref{fig:figure1} and analyze whether reducing the rank of $L$ damages performance. As Figure~\ref{fig:lowrank}c shows, reducing the rank of $L$ does not damage performance for tasks from Section~\ref{sec:arithmetic}, indicating that these transformations, in effect, project activations onto a lower-dimensional subspace.

\vspace{-1em}
\section{Conclusion}
We combine attention weights from previously-discovered components to obtain low-rank transformations that reveal token and concept information in Llama-2-7b, suggesting that understanding the geometry of LLM activations requires a precise formulation of \textit{what information} we want to analyze. 



\bibliographystyle{plain} 
\bibliography{references}


\newpage
\appendix

\section{Full Parallelogram Arithmetic Results}\label{app:full-nn}

\begin{table}[htbp]
\centering
\caption{Prefixes and examples for parallelogram datasets. Prefixes are used for all words in the dataset, e.g., ``She travelled to Athens'', ``She travelled to Greece'', etc.}
\label{tab:prefixes}
\begin{tabular}{@{}lll@{}}
\toprule
Task & Example & Prefix \\
\midrule
\multicolumn{3}{@{}l@{}}{\textbf{Word2Vec Tasks} (Mikolov et al., \cite{mikolov})} \\
\midrule
capital-common-countries & (Athens, Greece)  & She travelled to \\
capital-world & (Valletta, Malta) & She travelled to \\
currency & (Algeria, dinar) & You will have to pay in \\
city-in-state & (Tulsa, Oklahoma) & She travelled to \\
family & (uncle, aunt) & Did you talk to her \\
gram1-adjective-to-adverb & (amazing, amazingly) & Here is a random word in English: \\
gram2-opposite & (likely, unlikely) & Here is a random word in English: \\
gram3-comparative & (big, bigger) & Here is a random word in English: \\
gram4-superlative & (great, greatest) & Here is a random word in English: \\
gram5-present-participle & (look, looking) & Here is a random word in English: \\
gram6-nationality-adjective & (Brazil, Brazilian) & Here is a random word in English: \\
gram7-past-tense & (jumping, jumped) & Here is a random word in English: \\
gram8-plural & (cow, cows) & Here is a random word in English: \\
gram9-plural-verbs & (search, searches) & Here is a random word in English: \\
\midrule
\multicolumn{3}{@{}l@{}}{\textbf{Function Vector Tasks} (Todd et al., \cite{todd2024function})} \\
\midrule
antonym & (wish, regret) & Here is a random word in English: \\
synonym & (dangerous, hazardous) & Here is a random word in English: \\
present-past & (separate, separated) & Here is a random word in English: \\
singular-plural & (spoon, spoons) & Here is a random word in English: \\
word-length & (7, pelican) & Here is a random word in English: \\
capitalize-first-letter & (R, remember) & Here is a random word/character: \\
capitalize-last-letter & (T, quilt) & Here is a random word/character: \\
capitalize-second-letter & (N, snake) & Here is a random word/character: \\
lowercase-first-letter & (r, RACE) & Here is a random word/character: \\
lowercase-last-letter & (e, OBSERVE) & Here is a random word/character: \\
next-capital-letter & (ostrich, P) & Here is a random word/character: \\
next-item & (May, June) & Here is a random word/character: \\
prev-item & (twenty, nineteen) & Here is a random word/character: \\
capitalize & (peach, Peach) & Here is a random word in English: \\
country-capital & (Indonesia, Jakarta) & She travelled to \\
country-currency & (Slovenia, Euro (EUR)) & You will have to pay in \\
english-french & (discussed, discuté) & Voici un mot aléatoire en français: \\
english-german & (officials, Beamte) & Hier ist ein beliebiges Wort im Deutschen: \\
english-spanish & (forwards, adelante) & Aquí hay una palabra arbitraria en español: \\
landmark-country & (Chile, Wellington Island) & On vacation, we went to \\
national-parks & (California, Sequoia National Park) & On vacation, we went to \\
park-country & (Nepal, Bardya National Park) & On vacation, we went to \\
person-instrument & (piano, Tadd Dameron) & I am a big fan of \\
person-occupation & (architect, Gunnar Birkerts) & I am a big fan of \\
person-sport & (basketball, Kevin Durant) & I am a big fan of \\
product-company & (Apple, iPhone 5) & I am a big fan of \\
sentiment & (positive, It's a masterpiece.) & Here's my take on this film: \\
\bottomrule
\end{tabular}
\end{table}

\begin{figure}
    \centering
    \includegraphics[width=\linewidth]{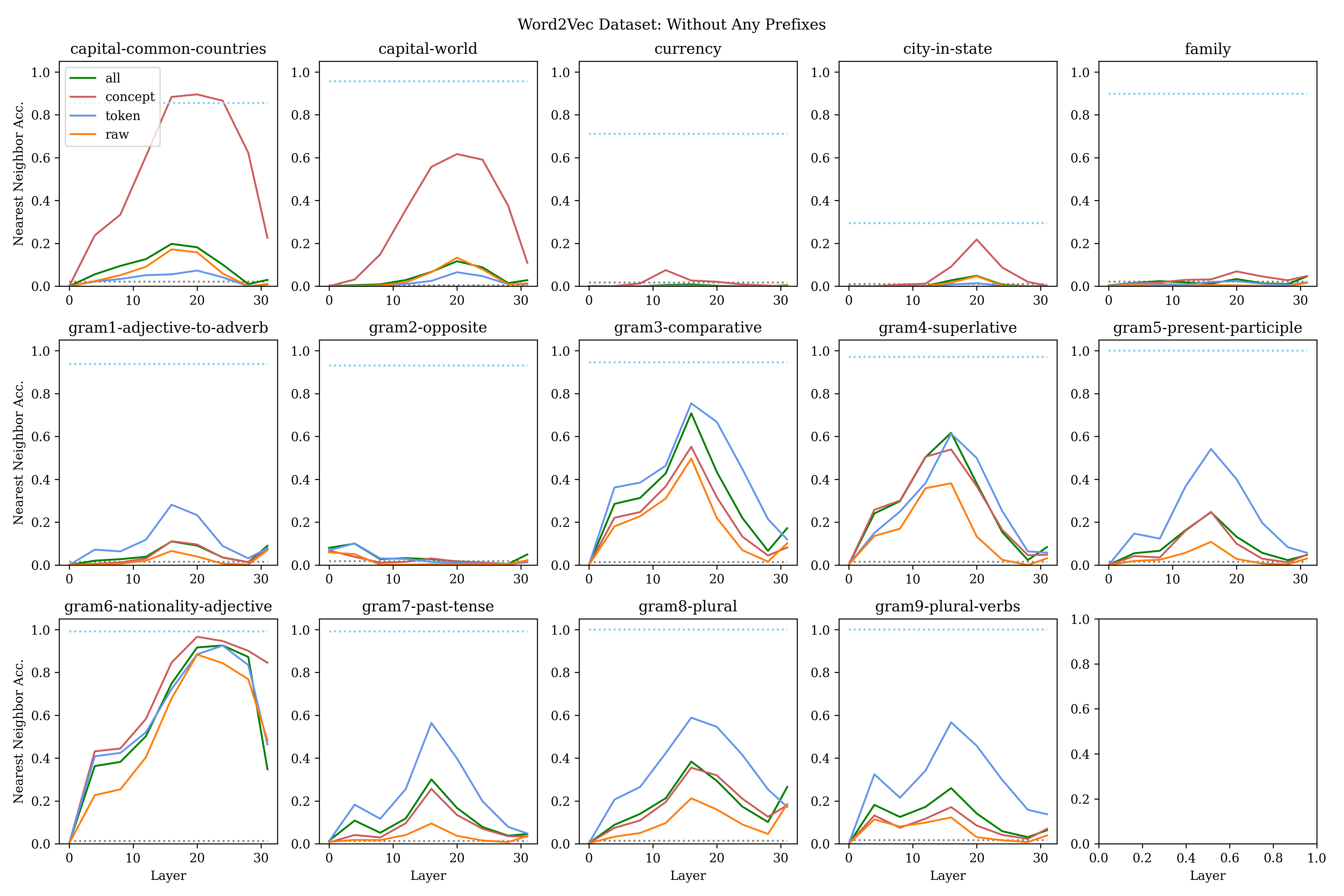}
    \caption{Nearest-neighbor accuracy for all \texttt{word2vec} tasks \citep{mikolov} without any prefixes (i.e., feeding each word to the model by itself with no context). Comparing with Figure~\ref{fig:word2vec-withprefix}, certain tasks like ``currency'' are much less accurate; this may be because currencies like ``real'' are not immediately recognizable out of context. However, accuracy is slightly better for ``capital-common-countries'' and ``gram6-nationality-adjective'' without any prefixes.}
    \label{fig:word2vec-noprefix}
\end{figure}

\begin{figure}
    \centering
    \includegraphics[width=\linewidth]{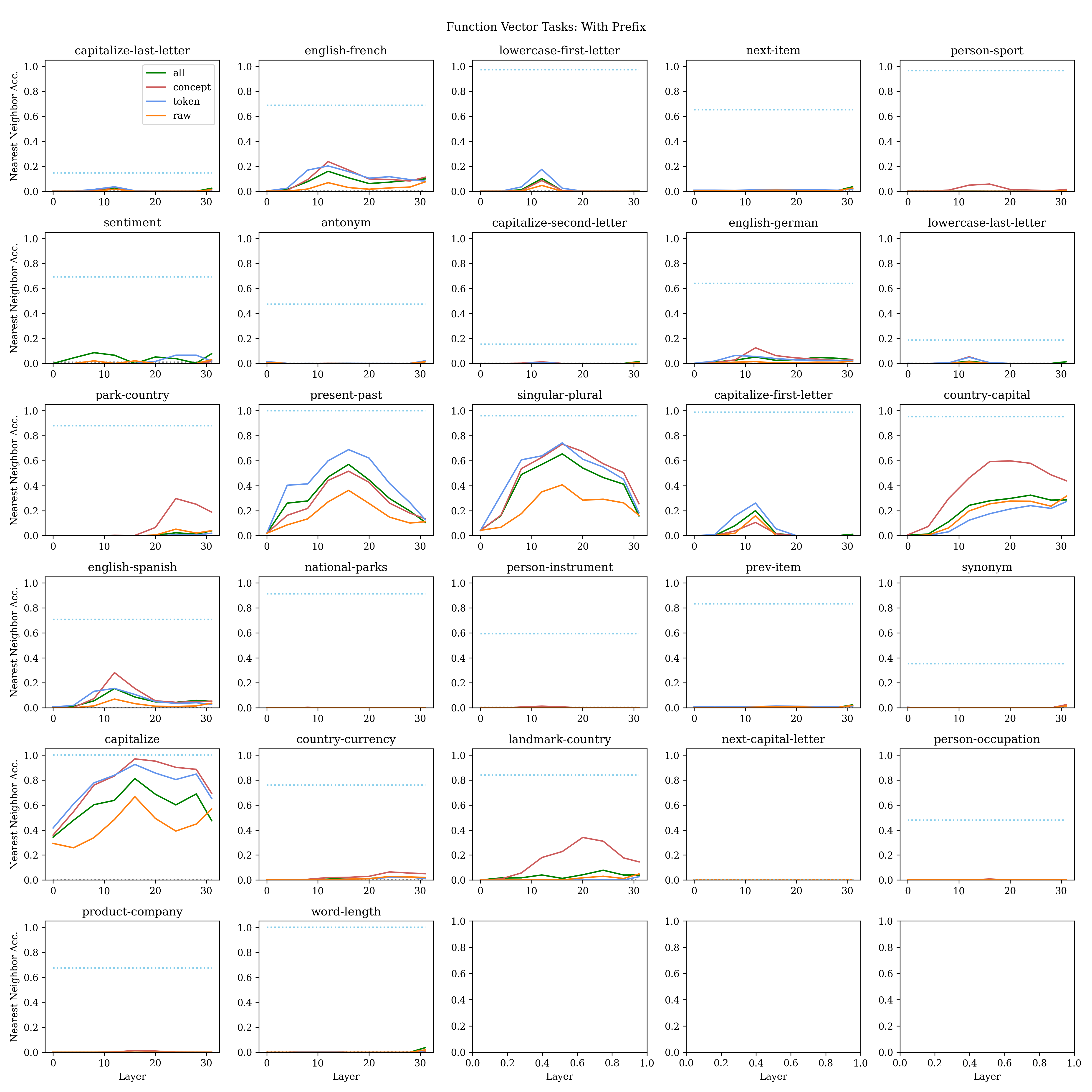}
    \caption{Nearest-neighbor accuracy for all function vector tasks \citep{todd2024function} with prefixes for each task listed in Table~\ref{tab:prefixes}. Dotted gray lines indicate guessing accuracy (out of all possible neighbors/words in the dataset). Dotted light blue lines indicate 5-shot ICL accuracy, i.e., the best possible performance this model can have for this task. The failure of many of these tasks is unsurprising: some tasks are many-to-one relations that may not be represented as parallelograms (``capitalize-first-letter''), whereas others may be too complex to be directly encoded in the model's embedding space (``national-parks''). Note: ``country-currency'' includes more countries (197) than the \texttt{word2vec} ``currency'' task (30).}
    \label{fig:fvs-withprefix}
\end{figure}

\begin{figure}
    \centering
    \includegraphics[width=\linewidth]{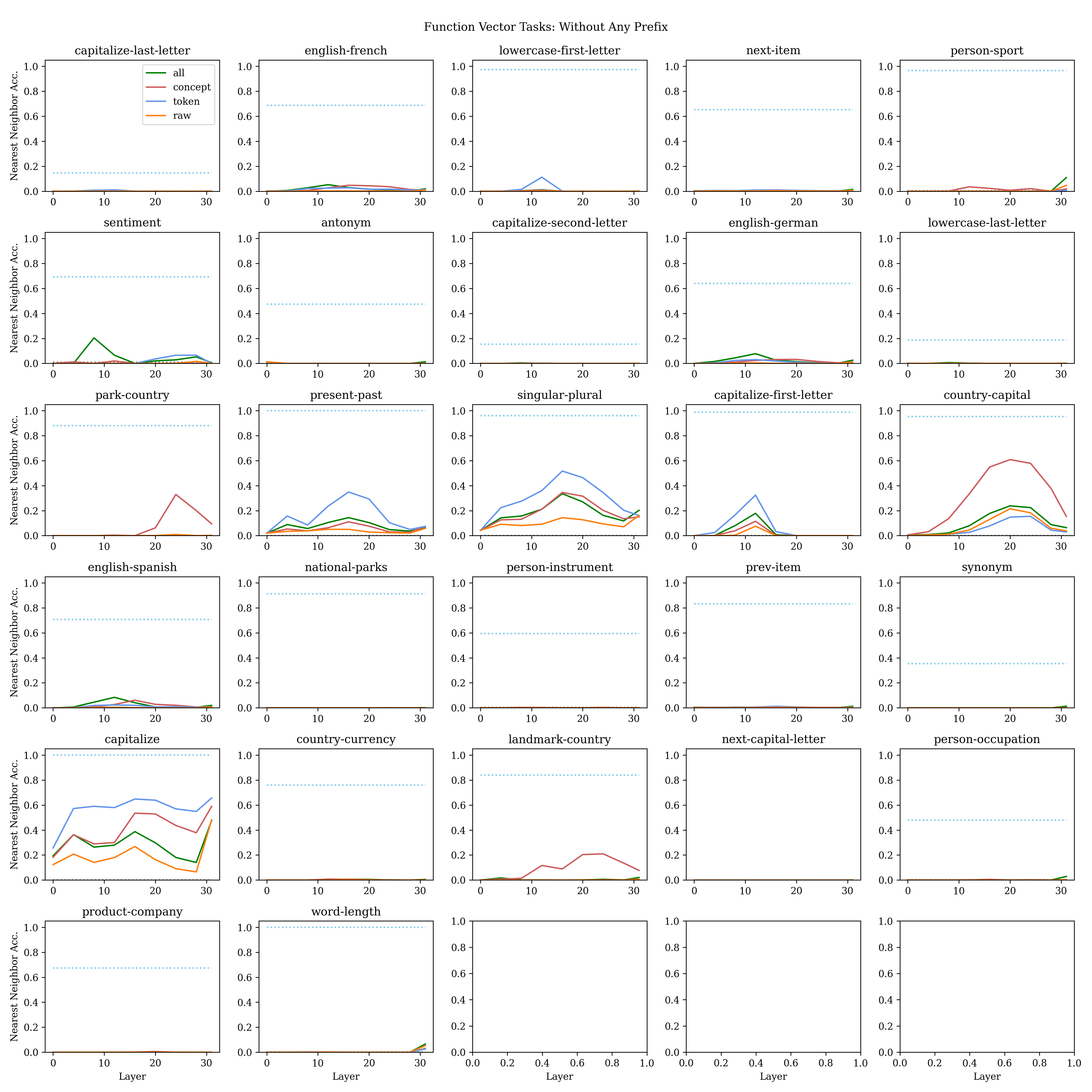}
    \caption{Nearest-neighbor accuracy for all function vector tasks \citep{todd2024function} without any prefixes (i.e., feeding each word to the model by itself with no context). Dotted gray lines indicate guessing accuracy (out of all possible neighbors/words in the dataset). Dotted light blue lines indicate 5-shot ICL accuracy for this task, i.e., the best possible performance this model can have for this task. Without prefixes, accuracy for many tasks is lower overall.}
    \label{fig:fvs-noprefix}
\end{figure}


\end{document}